%% file: main.tex
\documentclass[letterpaper]{article} 
\usepackage{aaai25}  
\usepackage{times}  
\usepackage{helvet}  
\usepackage{courier}  
\usepackage[hyphens]{url}  
\usepackage{graphicx} 
\usepackage{natbib}  
\usepackage{caption} 
\usepackage{algorithm}
\usepackage{algorithmic}

\usepackage[acronym,toc,xindy]{glossaries} 
\input{glossary_entries}

\usepackage{amsmath}
\usepackage{algorithm}
\usepackage{algorithmic}
\usepackage{booktabs}
\usepackage[table]{xcolor}
\usepackage{tabularx}
\usepackage{soul}
\usepackage{todonotes}
\usepackage{multirow}
\newcommand{\sep}{\,|\,}
\usepackage{bbding}
\usepackage{pifont}
\usepackage{wasysym}
\usepackage{amssymb}
\newcommand{\xmark}{\ding{55}}%

\usepackage{pgfplots}
\definecolor{blau}   {RGB}{  0  84 159}
\definecolor{grun}   {RGB}{ 87 171  39}
\definecolor{hellgrun}{RGB}{242 247 236}

\definecolor{hellrot}{RGB}{250 235 227}
\definecolor{bordeaux}   {RGB}{161  16  53}
\definecolor{rwth}   {RGB}{  0  84 159}
\definecolor{rwth-75}{RGB}{ 64 127 183}
\definecolor{rwth-50}{RGB}{142 186 229}
\definecolor{rot}   {RGB}{204   7  30}
\definecolor{rot-75}{RGB}{216  92  65}
\definecolor{rot-50}{RGB}{230 150 121}
\definecolor{grun}   {RGB}{ 87 171  39}
\definecolor{grun-75}{RGB}{141 192  96}
\definecolor{grun-50}{RGB}{184 214 152}

\tikzset{every picture/.style={/utils/exec={\sffamily}}}
\usepackage{subcaption}

\usepackage{newfloat}
\usepackage{listings}
\DeclareCaptionStyle{ruled}{labelfont=normalfont,labelsep=colon,strut=off} 
\floatstyle{ruled}
\newfloat{listing}{tb}{lst}{}
\floatname{listing}{Listing}
\title{LogicAD: Explainable Anomaly Detection via VLM-based Text Feature Extraction}
\author {
    Er Jin\textsuperscript{\rm 1}\textsuperscript{\rm,}\footnote{Corresponding authors with equal contributions.},
    Qihui Feng\textsuperscript{\rm 2}\textsuperscript{\rm,}\footnotemark[1],
    Yongli Mou\textsuperscript{\rm 2},
    Stefan Decker\textsuperscript{\rm 2}\textsuperscript{\rm,}\textsuperscript{\rm 3}, 
    Gerhard Lakemeyer\textsuperscript{\rm 2},\\
    Oliver Simons\textsuperscript{\rm 4}\textsuperscript{\rm,}\footnote{These authors contributed equally.}, 
    Johannes Stegmaier\textsuperscript{\rm 1}\textsuperscript{\rm,}\footnotemark[2]
}
\affiliations {
    \textsuperscript{\rm 1}Institute of Imaging and Computer Vision, RWTH Aachen University, Aachen, Germany\\
    \textsuperscript{\rm 2}Department of Computer Science, RWTH Aachen University, Aachen, Germany\\
    \textsuperscript{\rm 3} Fraunhofer Institute for Applied Information Technology FIT, Sankt Augustin, Germany
    \textsuperscript{\rm 4} Independent\\
    \{er.jin, johannes.stegmaier\}@lfb.rwth-aachen.de, \{feng, gerhard\}@kbsg.rwth-aachen.de,\\ \{mou, decker\}@dbis.rwth-aachen.de, science.osimons@posteo.de 
}

\usepackage{bibentry}

\begin{document}

\maketitle

\begin{abstract}
Logical image understanding involves interpreting and reasoning about the relationships and consistency within an image's visual content. This capability is essential in applications such as industrial inspection, where logical \gls{ad} is critical for maintaining high-quality standards and minimizing costly recalls. Previous research in \gls{ad} has relied on prior knowledge for designing algorithms, which often requires extensive manual annotation effort, significant computing power, and large amounts of data for training. \glspl{avlm} offer a promising alternative due to their exceptional performance in visual reasoning across various domains. Despite this, their application in logical \gls{ad} remains unexplored. In this work, we investigate using \glspl{avlm} for logical \gls{ad} and demonstrate that they are well-suited to the task. Combining \glspl{avlm} with format embedding and a logic reasoner, we achieve \gls{sota} \gls{ad} performance on public benchmarks, MVTec LOCO AD, with an AUROC of $\textbf{86.0\%}$ and an $F_1$-max of $\textbf{83.7\%}$ along with explanations of the anomalies. This significantly outperforms the existing \gls{sota} method by $\textbf{18.1\%}$ in AUROC and $\textbf{4.6\%}$ in $F_1$-max score. The dataset, code and supplementary materials are available at https://jasonjin34.github.io/logicad.github.io/.
\end{abstract}

\section{Introduction}

Anomalies in industrial image data can be broadly classified into two distinct categories: \glspl{sa} and \glspl{la}~\cite{DBLP:journals/ijcv/BergmannBFSS22}. The structures of \gls{sa} observed in industrial images are often referred to as localized regional features, such as broken parts, color contamination, and minor deformations. These anomalies are typically observable only in the abnormal object. In contrast, logical anomalies, such as missing objects, misplacements, and incorrect object color combinations, are not confined to a specific area. Generally, detecting them requires a more comprehensive, abstract understanding of the normal and abnormal states. Moreover, some LA features even appear in both abnormal and normal objects~\cite{Bergmann2019MVTecADComprehensive, Zou2022SPotDifferenceSelf}. Current methods face challenges in effectively detecting \gls{la}. The full-shot method, which is trained on a full dataset in the AD domain, captures all the error-free localized features as a memory bank and uses them to detect anomalies, which can be very effective at SA detection while facing difficulties in detecting \gls{la}s~\cite{roth2022towards, kim2024few}. 

\begin{figure}[t]
    \centering
    \includegraphics[width=\columnwidth]{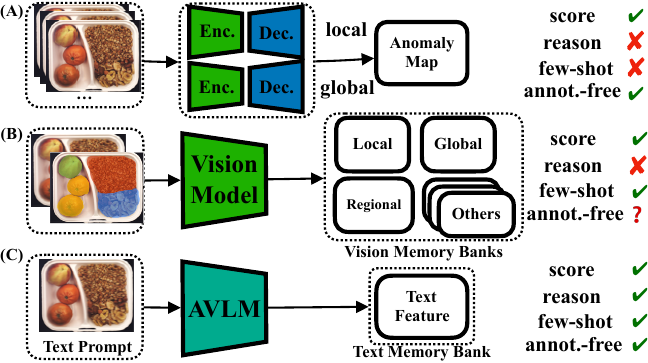}
    \caption{
\textbf{Overview of Anomaly Detection Approaches}:
(A) \gls{ad} models trained from scratch require large-scale datasets and are capable of \gls{ad} but lack reasoning capabilities. (B) Memory-based \gls{ad} methods leverage a pre-trained vision model to extract features from normal images, enabling \gls{fs} \gls{ad}. However, they often require additional visual annotations and lack reasoning. (C) Our method uses pre-trained \glspl{avlm} as a text feature extractors and uses it for \gls{la} detection and reasoning with only text prompts, eliminating the need for visual annotations. 
}
    \label{fig:approach} 
\end{figure}

As shown in Figure~\ref{fig:approach}(A), some approaches attempt to solve this issue by training multiple global and local \gls{ae} networks to capture both \gls{la} and \gls{sa} related features~\cite{DBLP:journals/ijcv/BergmannBFSS22}. Others rely on additional manual visual annotations and multiple memory banks to achieve remarkable \gls{la} detection~\cite{kim2024few, zhang2024logicode}. However, the inability to perform few-shot learning and the requirement for additional manual visual annotation is undesirable. Recently, many methods based on vision-language models, such as \gls{clip}~\cite{radford2021learning}, have shown remarkable performance in \gls{fs} \gls{sa} detection~\cite{Jeong2023WinCLIPZero/Few, chen2023april}. In \gls{la} detection, particularly under \gls{fs} learning scenarios, the challenge of understanding long-range global features remains unresolved. Recently, many methods, as shown in Figure~\ref{fig:approach}(B), aim to capture global features by using multiple memory banks using \gls{moe}~\cite{admoe}. However, compared with other full-shot methods, which are trained with the whole dataset, the performance is noticeably inferior~\cite{kim2024few}. Figure \ref{fig:approach} provides an overview of current \gls{la} detection methods (A) and (B), which are predominantly based on visual features. Recent advancements in \glspl{avlm} such as GPT-4o and LLaVA 1.6 \cite{achiam2023gpt, liu2024visual} have demonstrated significant capabilities in image understanding and text generation. Despite these advancements, using text features for \gls{ad}, particularly logical anomalies, remains underexplored. In this paper, we introduce our \gls{ad} algorithm, LogicAD, which primarily utilizes text features extracted from \glspl{avlm} rather than relying solely on visual features and without relying on additional visual annotations. Our evaluations on multiple public datasets reveal that LogicAD surpasses the current \gls{sota} method by a large margin. Our contributions are as follows: \textbf{(1)} We introduce LogicAD, a novel one-shot algorithm for \gls{la} detection that leverages text feature memory banks, employing \glspl{avlm} and \gls{llms} to achieve the \gls{sota} performance in one-shot logical \gls{ad}. \textbf{(2)} We design a text feature extraction pipeline that enables \glspl{avlm} to generate logical, robust, and reliable text features for detailed logical descriptions. \textbf{(3)} We introduce a \textit{logic reasoner}, which leverages \gls{atp} for \gls{la} detection and  generates descriptive explanations for the identified anomaly without using manual or dynamic thresholding.

\begin{figure*}[t]
    \centering
    \includegraphics[width=\textwidth]{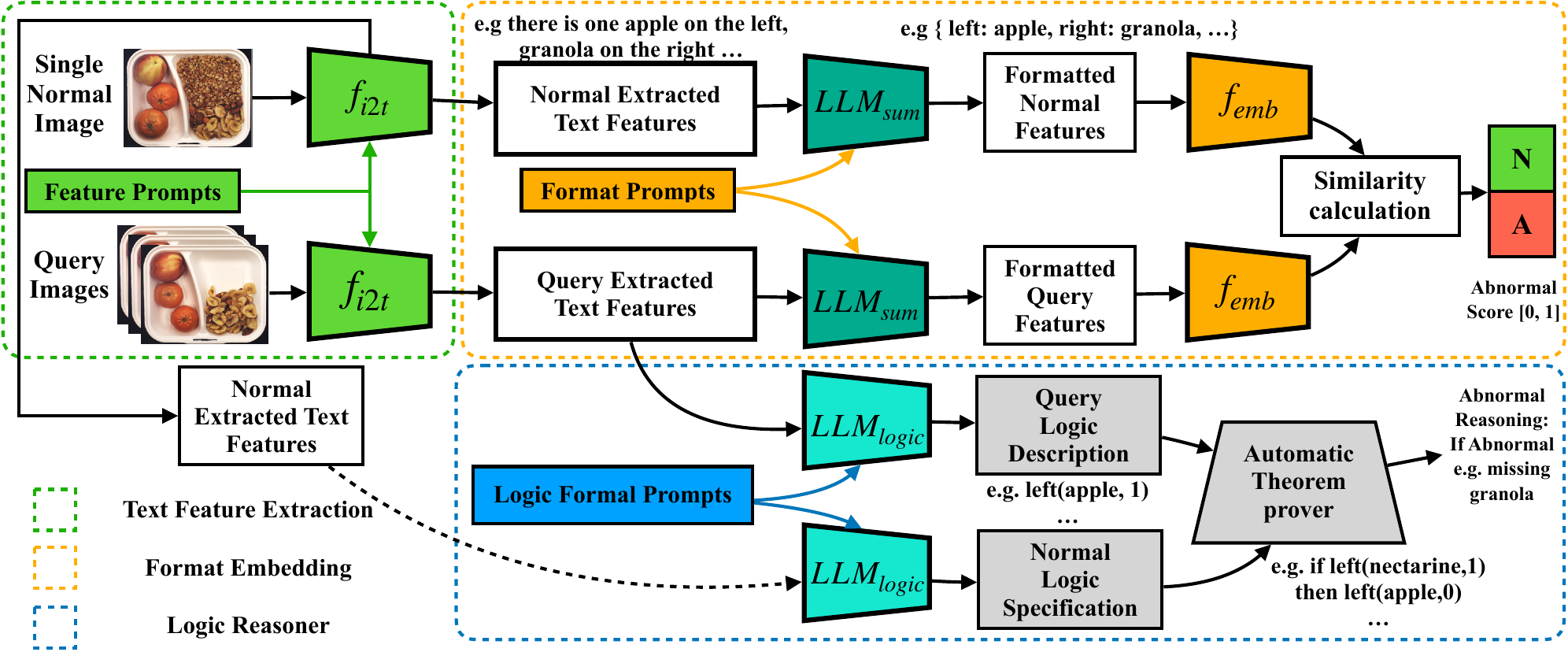}
    \caption{\textbf{Pipeline overview of LogicAD.} The \textcolor{green}{green box} represents \textit{text feature extraction}, $f_{i2t}$, which extracts features via pre-trained \glspl{avlm} from the input image, the detailed process is depicted in Figure~\ref{fig:text_extraction}. These features are then processed by two separate modules: \textit{format embedding} (\textcolor{orange}{orange box}) and \textit{logic reasoner} (\textcolor{blue}{blue box}). The \textit{format embedding} module computes an anomaly score based on the similarity between embeddings of formatted normal and query features. The \textit{logic reasoner} module utilizes logical rules derived from normal data to classify inputs as normal or abnormal while providing reasoning.}
    \label{fig:main_figure} 
\end{figure*}

\section{Related Work}

\noindent \textbf{\Gls{la} in Industrial Images.} Following the release of the logical anomaly dataset MVTec LOCO AD \cite{DBLP:journals/ijcv/BergmannBFSS22}, numerous unsupervised methods have been proposed~\cite{liu2023component, rudolph2023asymmetric}. These methods can be categorized into vision memory bank-based methods and reconstruction networks. As illustrated in Figure~\ref{fig:approach}(A), GCAD employs multiple high-capacity \gls{ae} networks to capture the global context through its latent space. For SA, GCAD requires an additional local branch. Furthermore, each branch demands a significant amount of data to learn and capture the latent features, posing a limitation in few-shot scenarios \cite{DBLP:journals/ijcv/BergmannBFSS22}. Inspired by PatchCore~\cite{roth2022towards}, vision memory-based methods have gained popularity in the few-shot domain due to their simplicity and effectiveness. Visual features are often extracted via pre-trained networks, commonly using CNN-based architectures such as ResNet~\cite{olivermdnpd, rippel2021gaussian, liu2023component, zhouanomalyclip, he2016deep}. To understand logical-related long-range contexts, ComAD~\cite{liu2023component} utilizes DINO~\cite{caron2021emerging} for segmenting images into \glspl{roi} and extracting region-based features as a memory bank. PSTD~\cite{kim2024few} proposes using a few fully-annotated segmentation masks to help models capture long-range contexts. However, PSTD still requires three vision-based memory banks. ViperGPT utilizes the \gls{avlm} to generate task-related code for handling downstream tasks such as \gls{ad}~\cite{suris2023vipergpt}. However, framing \gls{ad} as a vision-centric task is challenging due to the high level of semantic understanding required. Additionally, the generated code can introduce bias and increase computational cost. AnomalyGPT proposed an approach that fine-tunes open source \gls{llms}~\cite{touvron2023llama} and a vision encoder for effective \gls{sa} detection and even achieve anomaly localization without the need to use a manual threshold. Another similar method, VisionLLM, can be adapted for \gls{ad} \cite{wang2024visionllm}.  However, these methods require synthesized anomalous datasets and demand substantial computational resources for fine-tuning, making the process resource-intensive and computationally demanding. In contrast, our \gls{fs} method (one-shot) can achieve \gls{ad} without any fine-tuning. It is also the first method to integrate \gls{avlm} capabilities with a purely logical reasoning framework,  which is  further supported by an automated theorem prover, significantly improving the explainability of anomaly detection results.

\noindent\glsreset{avlm}\textbf{\glspl{avlm}.} Recently, many \glspl{avlm}, such as GPT-4o and LLaVA1.6~\cite{achiam2023gpt, liu2024visual}, have achieved remarkable results across multiple benchmarks, such as VQA-v2~\cite{goyal2017making} and ScienceQA~\cite{lu2022learn}. These models perform exceptionally well in tasks involving naive logical scenarios, such as object localization and size comparisons. However, in real-world scenarios, researchers have noted that \glspl{avlm} often struggle with hallucinations, where the models fail to accurately ground both the provided text and visual context~\cite{DBLP:conf/aaai/GunjalYB24}. For instance, the ability of \glspl{avlm} to understand quantitative and logical information decreases as the complexity of the data increases, as demonstrated in Figure~\ref{fig:gpt-count}. Research suggests that using \gls{cot} can partially mitigate the hallucination issue in \glspl{avlm}~\cite{DBLP:conf/aaai/ChenZSHSGG24}. However, the effectiveness of \gls{cot} in industrial \gls{ad} has not yet been fully explored.

\noindent \textbf{Logic Reasoning.} The development of \gls{llms} also boosts the investigation of neuro-symbolic approaches, which combine language models with formal methods by parsing natural language statements into formal languages such as first-order logic~\cite{enderton2001mathematical}. Previous work such as LINC~\cite{olausson2023linc}, Logic-LM~\cite{pan-etal-2023-logic} and SatLM~\cite{ye2024satlm} utilize \gls{llms} as a semantic translator and convert natural language problems into formal specifications. Then, \gls{atp} such as Prover9\footnote{https://www.cs.unm.edu/~mccune/prover9/} performs inference to solve the queries. For tasks with complex logical relations, these approaches significantly outperform in-context reasoning methods such as \gls{cot}, inspiring us to combine our system with formal methods to handle complicated relations among objects and to explain detection results.

\section{LogicAD}

\textbf{Problem Definition:}  \Gls{ad} is a task that focuses on identifying abnormal features by learning from normal features denoted as $F_{normal}$ extracted from a set of training images ${\{X_1, X_2, ..., X_N\}}$, where $N$ denotes the total number of anomaly-free training images. \Gls{zs} \gls{ad} leverages pre-trained \gls{vlm} with provided text prompts for \gls{ad} without any training images. \Gls{fs} uses few images or even one training image (One-shot) for \gls{ad}. Compared to the \gls{zs} approach, which relies solely on reference text descriptions, the \gls{fs} method uses \glspl{avlm} to generate reference descriptions, simplifying prompt creation while making the process considerably more \gls{avlm}-agnostic. 

\noindent \textbf{Overview:} We propose the LogicAD algorithm, which consists of three primary components: \textit{text feature extraction}, \textit{format embedding} and \textit{logic reasoner}, as shown in Figure \ref{fig:main_figure}. The \textit{text feature extraction}, $f_{i2t}$, which converts image to text, generates consistent and reliable logical text descriptions. The \textit{format embedding} component calculates an anomaly score to identify deviations from normal patterns. Meanwhile, the \textit{logic reasoner} generates textual reasonings for the observed anomalies. Together, these components enable not only \gls{ad} but also the explanation of the underlying reasons.

\subsection{Text Feature Extraction}

\begin{figure}[h!]
    \centering
    \includegraphics[width=\columnwidth]{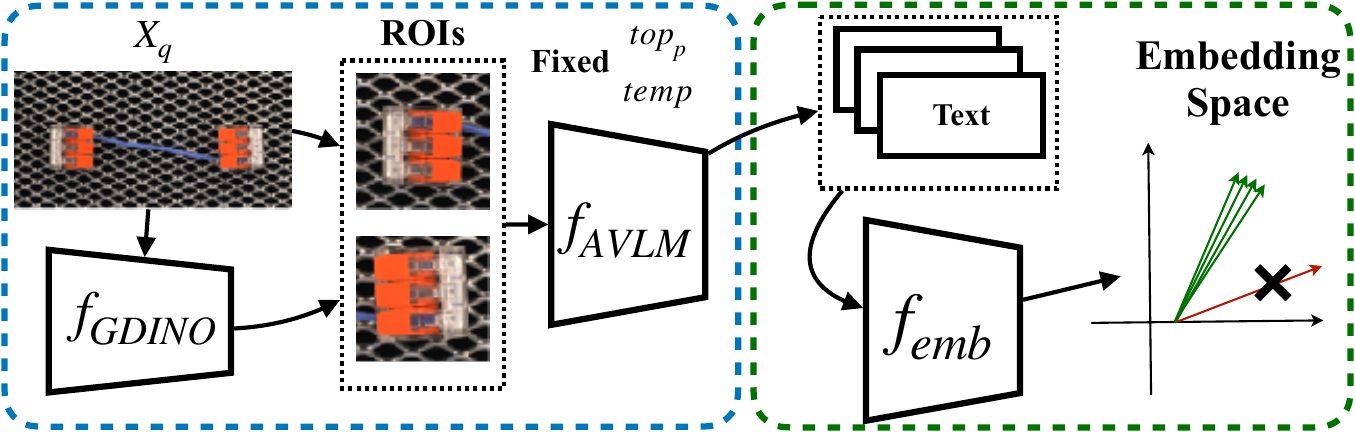}
    \caption{\textbf{Text feature extraction} $f_{i2t}$ involves \gls{roi} extraction (blue box) and text embedding filtering (green box). Patches and the original image are processed by an \gls{avlm} to generate $K$ text descriptions. The green box uses text-embedding-3-large \cite{achiam2023gpt} for output stabilization.    
    }
    \label{fig:text_extraction} 
\end{figure}

\begin{figure}[h!]
    \centering
    \includegraphics[width=\columnwidth]{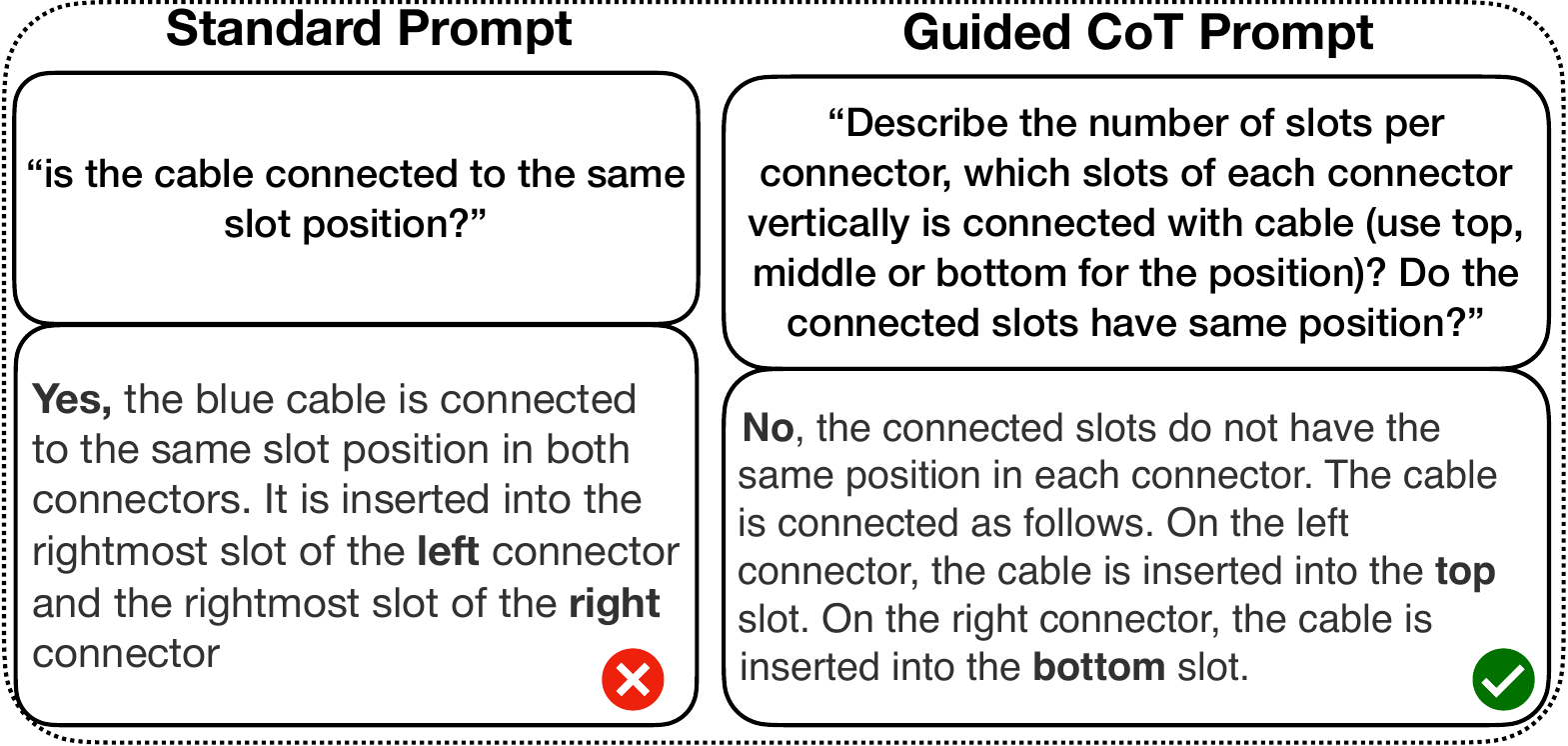}
    \caption{\textbf{Illustration of a standard prompt versus a \gls{gcot} prompt.} We use image $X_q$ from Figure~\ref{fig:text_extraction} as the input. The ground truth description specifies that two cables are not connected to the same slot position. Using prompts based on \gls{gcot}, \glspl{avlm} can generate more accurate descriptions of the input image.}
    \label{fig:cot} 
\end{figure}

Vision-language models such as CLIP \cite{radford2021learning}, ImageBind~\cite{girdhar2023imagebind}, and EVA-CLIP~\cite{sun2023eva} are extensively utilized in zero-shot and \gls{ad} algorithms~\cite{Jeong2023WinCLIPZero/Few, gu2024anomalygpt}. However, in the domain of \gls{la} detection, \gls{clip}-based VLMs exhibit significant limitations, yielding markedly poorer performance on tasks involving logical inconsistencies compared to their effectiveness in detecting \gls{sa}. Table~\ref{tab:mvtec_eval} (categories shaded in grey) shows that WinCLIP performs significantly worse in categories containing primarily naive logic-related anomalies, such as missing or mislocated objects~\cite{Jeong2023WinCLIPZero/Few}.

\noindent\textbf{\Glspl{avlm}}, such as BLIPv2, LLaVA and GPT-4o~\cite{li2023blip, liu2024visual, achiam2023gpt} have demonstrated remarkable potential in image understanding but also face challenges in handling logic-related tasks such as counting objects, localizing objects, understanding chains of logic~\cite{pan-etal-2023-logic} along with inconsistent results \cite{pan-etal-2023-logic, achiam2023gpt}. To alleviate these issues, we designed our \textit{text feature extraction} module as shown in Figure \ref{fig:text_extraction} with the following components: \gls{gcot}, \gls{roi} segmentation, and text embedding filtering. 

\noindent\textbf{\gls{gcot}.} WinCLIP proposes the Compositional Prompt Ensemble (CPE), which generates all predefined text prompts containing words that describe the state of the object, such as "flawless, damaged, or defect" ~\cite{Jeong2023WinCLIPZero/Few}. Many other competing methods ~\cite{zhouanomalyclip, gu2024anomalygpt} are based on WinCLIP. CPE is suitable for \gls{sa} but is insufficient for handling \gls{la}, as shown in Table \ref{tab:mvtec_eval}. Inspired by \cite{wei2022chain}, we propose to use a \gls{gcot} for logical description. We guided the \glspl{avlm} model to inspect the images based on \gls{gcot} text prompts, as shown in Figure \ref{fig:cot}. Without providing specific location-related logic guides, such as describing the position only "vertically", or numerical logic guides, such as "specifying the number of slots per connector", \glspl{avlm} such as GPT-4o fail to comprehend the semantic meaning of localization, leading to logical hallucinations. Appendix A.1 demonstrates more \gls{gcot} examples. Although our method requires manual prompt creation, providing detailed and reusable text prompts is significantly easier in practical applications than other methods requiring visual annotations~\cite{kim2024few, zhang2024good} (Figure~\ref{fig:approach}). 

\noindent \textbf{\gls{roi} Segmentation and Text Embedding Filtering.} Current autoregressive Vision-Language Models have limitations in logical reasoning tasks, such as counting, object-size estimation, localization, and basic calculations \cite{achiam2023gpt, zhang2024good, lee2023mathematical}. To further investigate these limitations, we evaluated GPT-4o using a subset of the CountBench dataset \cite{paiss2023teaching}. We randomly selected 250 images from this dataset. Additionally, we created a custom dataset, UniformBench composed of 150 images characterized by homogeneous features such as varying numbers of pawns on a chessboard, beer bottles in a basket, and Go boards with random distributions of white and black stones. More sample images from CountBench and UniformBench are shown in Appendix A.2. As illustrated in Figure \ref{fig:gpt-count}, GPT-4o demonstrates high accuracy in object counting tasks within CountBench, where images often present heterogeneous features. However, the model's performance significantly declines when tested on the UniformBench dataset, which lacks these additional heterogeneous features. This reduction in accuracy correlates with increasing task complexity in object counting, demonstrating a critical limitation in the model's ability to handle essential tasks for industrial logical \gls{ad}.

\begin{figure}[t]
    \centering
    \begin{subfigure}[t]{0.62\columnwidth} 
        \centering
        \resizebox{!}{0.7\linewidth}{
        \begin{tikzpicture}
        \begin{axis}[
            width=13cm,
            height=9cm,
            ymin=0, ymax=110,
            ybar=0pt,
            bar width=0.47cm,
            enlarge x limits=0.05, 
            ylabel={\textbf{\LARGE Accuracy \%}},
            xlabel={\textbf{\LARGE Number of Objects}},
            symbolic x coords={1, 2, 3, 4, 5, 6,7, 8, 9, 10},
            xtick=data,
            legend pos=south west,
            legend cell align={left},
            nodes near coords,
            every node near coord/.append style={font=\LARGE},
            axis line style={-}, 
            tick style={-}, 
            label style={font=\LARGE}, 
            tick label style={font=\LARGE}, 
            legend style={font=\LARGE}, 
        ]
        \addplot[draw opacity=0,fill=rwth-75,text=rwth] 
            coordinates {(1,90) (2, 95) (3,74) (4,85) (5,80) (6,80) (7,90) (8,100) (9,90) (10,85)};
        \addplot[draw opacity=0,fill=rot-75, text=rot] 
            coordinates {(1,100) (2,57) (3,80) (4,87) (5,83) (6,42) (7,52) (8,45) (9,35) (10,36)};
        \legend{CountBench,UniformBench}
        \end{axis}
        \end{tikzpicture}
        }
        \caption{\textbf{The accuracy of GPT-4o in counting.}}
        \label{fig:gpt-count}
    \end{subfigure}
    \hfill
    \begin{subfigure}[t]{0.35\columnwidth} 
        \centering
        \includegraphics[width=0.75\columnwidth]{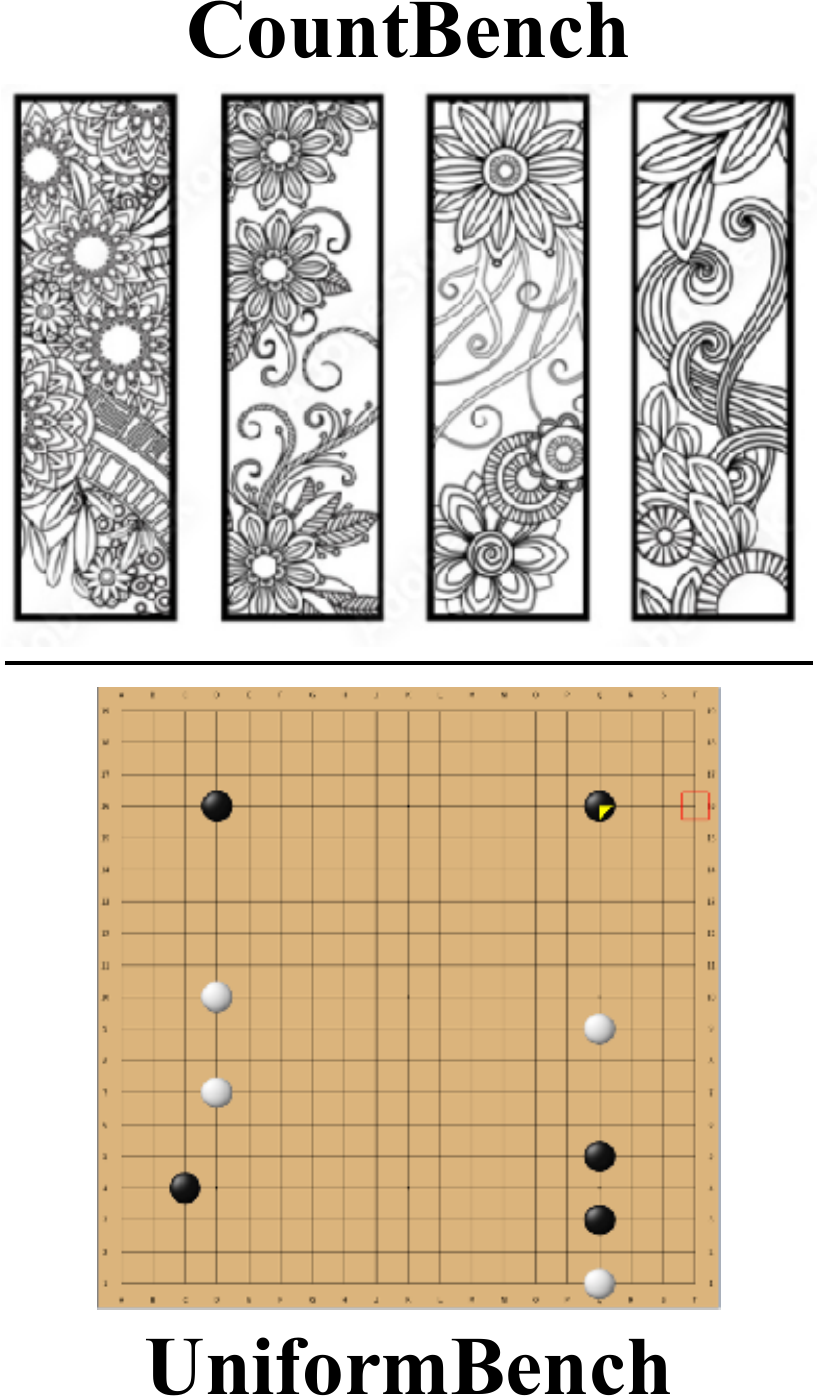} 
        \caption{\textbf{Sample Images}}
        \label{fig:ds}
    \end{subfigure}
    \caption{\textbf{Comparison of GPT-4o counting accuracy and additional visual examples.} Accuracy of GPT-4o drops significantly with an increasing number of homogeneous objects. Figure~\ref{fig:ds} shows two samples, one from CountBench (top) and one from UniformBench (bottom).}
    \label{fig:combined}
\end{figure}
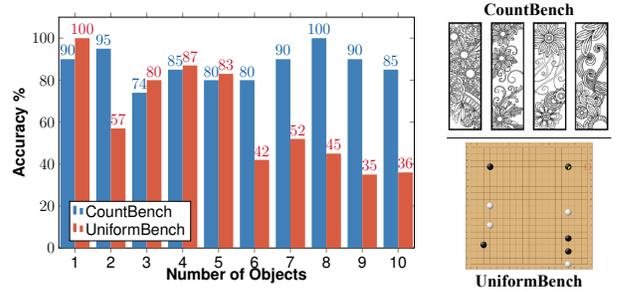
To alleviate this issue, we propose reducing the complexity by selecting \gls{roi}s from the image, using GroundingDINO as our \gls{roi} extraction model \cite{liu2023grounding}. All the prompts used for GroundingDINO are keywords directly extracted from \glspl{avlm} without any additional effort. The detailed prompts are provided in Appendix A.1. Most \glspl{avlm} demonstrate a noticeable level of inconsistency even when using the same seed or tuning the hyperparameters of \textit{$top_p$} and \textit{temperature} \cite{song2024good}. We suggest generating the extracted text multiple times and then using an embedding model as a filter to eliminate outlier text by using the Local Outlier Factor (LOF) method \cite{breunig2000lof}. The detailed steps of text extraction can be summarized as follows: \textbf{(1)} Generating a set of regions ${\mathbf{w}}_{i}$, where $i \in [1, N]$ and $N$ represents the total number of \glspl{roi}, using the function $f_{GDINO}$ with feature prompts. Each region, along with the original image, is then processed by the function $f_{AVLM}$, $K$ = 3 times, yielding a collection of textual descriptions $\mathcal{T} = \{t_1, t_2, t_3\}$. \textbf{(2)} Constructing the text embedding space $\mathcal{M}$ by applying the text embedding model, text-embedding-3-large from OpenAI \cite{achiam2023gpt},  $f_{emb}$ to $\mathcal{T}$, resulting in $\mathcal{M} = f_{emb}(\mathcal{T}) = \{\mathbf{e}_i\}_{i=1}^{k}$, where $\mathbf{e}_i$ is the embedding of the extracted text. Subsequently, these text embeddings are fed into an outlier detection model, specifically the Local Outlier Factor (LOF) function $f_{LOF}$ , to generate the filtered text embedding space, which is denoted as $\mathcal{T}_{filter}$.  We then randomly select  corresponding text from the filtered embedding space $\mathcal{T}_{filter}$.

\subsection{Format Embedding}
After extracting the text features, we use an LLM to summarize the text into \textit{JSON} format. Both the normal/reference image \(X_{n}\) and the query image \(X_{q}\) are then fed into the embedding function \(f_{emb}\) to generate the respective embedding features $\mathbf{\hat{e}}_n$ and $\mathbf{\hat{e}}_q$. We then calculate the anomaly score based on the cosine similarity between these normalized embeddings as follows: $ ascore = 1 - \langle \mathbf{\hat{e}}_n, \mathbf{\hat{e}}_q \rangle$. 

\subsection{Logic Reasoner}
We aim not only to achieve high performance in \gls{ad}, but also to make the detection results explainable. To achieve the latter, we convert the text summary into a formal specification and use an external automated theorem prover (ATP) to perform logical reasoning. To obtain formal specifications of an image, we use an LLM and a two-shot prompt. 
Consider the category ``breakfast box'' from MVTec LOCO AD~\cite{DBLP:journals/ijcv/BergmannBFSS22} for example, the hypothetical text features of an image include the following statement:

\textit{``On the left side of the box, there is a nectarine and an apple, and on the right, there are some nuts...''}\\
The formal specification (denoted as $\Sigma_0$) generated via LLM is as follows:
\begin{equation*}
left(nectarine, 1) \text{; } left(apple,1) \text{; } right(nut,irrel)\ldots
\end{equation*}
Here $irrel$ means that the number (of the nuts) is irrelevant.
Identifying anomalies via logical reasoning requires a rigorous specification of what is normal. Consider a description of normality as follows (It may not necessarily be the definition of normality in the original dataset. Our approach can be easily adopted for different definitions of ``anomaly''):

\textit{``On the left side of the plate, either there is an apple and no nectarine, or there is a nectarine but no apples. Besides, there should be two tangerines on the left...''}\\
Based on that, we derive the normal specification $\Sigma_{norm}$:
\begin{align*}
    &\big((left(apple, 1) \land left(nectarine,0)) \vee \\
    & \hspace{1cm} (left(nectarine,1) \land left(apple,0))\big)\\
    &left(tangerine,2) \ldots
\end{align*}
$\Sigma_{norm}$ can include any relational formulae. We exclude function symbols since empirically, they reduce the stability and precision of parsing. Due to the ambiguity of natural languages, some features may not be covered here: When one says "there is \textbf{an} apple on the left", implicitly, we rule out other numbers. For things that do not occur in the image, one may not explicitly specify their numbers. Hence, we use the following formulae to complete the logic program:

\begin{itemize}
    \item $\Sigma_{na}$: Inspired by the \emph{unique name assumption}, we use a set of (in-)equalities to explicitly specify whether two constants denote the same object or not, e.g.
    \begin{equation*}
        tangerine = mandarin\text{ , } tangerine \neq apple \text{ , } \ldots
    \end{equation*}
    The distinguishability of constants is obtained by the response to LLM queries such as ``\textit{Answer Yes or No: Are $object1$ and $object2$ synonymous or similar?}'' 
    \item $\Sigma_{fa}$: We can specify if a predicate has functional behaviour. For instance, the predicate $right$ denotes the number of instances of an object on the right, and each object should be assigned a unique number Thus $\Sigma_{fa}$ will include:
    \begin{equation*}
        \forall x \exists y. right(x,y) \land (\forall y'. right(x,y')\to (y'=y))
    \end{equation*}
    One can manually specify if a predicate is functional, or it can be automatically decided by an LLM query. Appendix A.3 shows some detailed examples for multiple categories.
    \item $\Sigma_{dca}$: To identify out-of-domain anomalies such as ``there's a bug on the plate'', we introduce axioms which serve as the \emph{domain closure}~\cite{reiter1980equality}:
    \begin{equation*}
        \forall x. (\neg left(x,0) \to (x=apple \sep x = tangerine \sep \cdots))
    \end{equation*}
    On the right-hand side, this is a disjunction of all constants mentioned in the normal specification $\Sigma_{norm}$. With this axiom, unmentioned objects should not exist on the left, i.e. $left(bug,1)$ will be verified as an anomaly. 
    \item Default: Conversely, to identify the missing item, we complete the formal description by adding default values: for example, if $left(tangerine,\_\_)$ (or anything synonymous with tangerine) is not mentioned in the image description $\Sigma_0$, then $left(tangerine,0)$ is added to $\Sigma_0$.
\end{itemize}

Among the aforementioned formulae, $\Sigma_{norm}$ and default values are essential and need to be provided for each class of AD task (``breakfast box'', ``juice bottle'', etc.). The rest can be automatically generated or manually specified. 

Let $\Gamma = \Sigma_{norm} \cup \Sigma_{na} \cup \Sigma_{fa} \cup \Sigma_{dca}$ be the union of all hypotheses. The \gls{ad} is then converted to a task of theorem proving:
\begin{itemize}
    \item If $\Gamma \models \neg \Sigma_0$, then we label the image as abnormal.
    \item If $\Gamma \not\models \neg \Sigma_0$, then we label the image as normal
\end{itemize}
We use Prover9 for theorem proving. Here $\Gamma \models \neg \Sigma_0$ means $\Gamma$ logically entails $\neg \Sigma_0$, i.e. every logical model satisfying $\Gamma$ will also satisfy the negation of $\Sigma_0$. Since $\Sigma_0$ is the formal description of the image, it shows that the image description contradicts the normal cases and hence the image is abnormal. To identify the actual anomaly, we look for a minimal subset of $\Sigma_0$ which causes the anomaly, i.e. for $\Sigma_a \subseteq \Sigma_0$, if $\Gamma \models \neg \Sigma_a$ and for any $\Sigma' \subsetneq \Sigma_a$, we have $\Gamma \not\models \neg \Sigma'$, then $\Sigma_a$ forms an explanation: Consider $\Sigma_0$ and  $\Gamma$ defined as above with all the mentioned formulae, then $\Gamma \models \neg \Sigma_0$ since both an apple and a nectarine are on the left. Then $\Sigma_a= \{left(nectarine,1), left(apple,1)\}$
is a formal explanation of the anomaly since $\Sigma_a \subseteq \Sigma_0$, and 
\begin{align*}
    & \Gamma \models \neg(left(nectarine,1)\land left(apple,1))\\
    & \Gamma \not\models \neg left(nectarine,1)\\
    & \Gamma \not\models \neg left(apple,1)
\end{align*}

\begin{table*}[h!]
\resizebox{\textwidth}{!}{
\begin{tabular}{@{}ccccccc
>{\columncolor[HTML]{EFEFEF}}c 
>{\columncolor[HTML]{EFEFEF}}c
>{\columncolor[HTML]{EFEFEF}}c 
>{\columncolor[HTML]{EFEFEF}}c @{}}
\toprule
\begin{tabular}[c]{@{}c@{}}MVTec LOCO AD \\ (only LA)\end{tabular}    & \multicolumn{2}{c}{\textbf{LogicAD} (Ours)}                            & \multicolumn{2}{c}{\begin{tabular}[c]{@{}c@{}} \textbf{AnomalyMoE}\textdagger\\ (CVPR VAND 24)\end{tabular}}   & \multicolumn{2}{c}{\begin{tabular}[c]{@{}c@{}} \textbf{WinCLIP} \\ (CVPR 23)\end{tabular}  } & \begin{tabular}[c]{@{}c@{}} \textbf{GCAD} \\ (IJCV 23)\end{tabular}  & \begin{tabular}[c]{@{}c@{}}PatchCore \\ (CVPR 22)\end{tabular}   & \begin{tabular}[c]{@{}c@{}}ComAD\\ (AEI 22)\end{tabular}  & \begin{tabular}[c]{@{}c@{}}AST\\ (ICCV 23) \end{tabular}  \\ \midrule
Category                        & AUROC                       & $F_1$-max                              & AUROC           & $F_1$-max                      & AUROC       &    $F_1$-max        & AUROC     & AUROC & AUROC & AUROC   \\ \midrule
Breakfast Box                   & 93.1$\pm$\footnotesize{2.1}              & 82.7$\pm$\footnotesize{1.4}                 & -               & -                                     &   57.6      &     63.3        & 87.0      & 74.8  & 94.5  &   80.0  \\
Juice Bottle                    & 81.6$\pm$\footnotesize{3.5}              & 83.2$\pm$\footnotesize{4.3}                 & -               & -                                     &   75.1      &     58.2        & 100.0       & 93.9  & 90.9  &   91.6  \\
Pushpins                        & 98.1$\pm$\footnotesize{0.1}             & 98.5$\pm$\footnotesize{0.1}                  & -               & -                                     &   54.9      &     57.3        & 97.5      & 63.6  & 89.0  &   65.1  \\
Screw Bag                       & 83.8$\pm$\footnotesize{5.2}             & 77.9$\pm$\footnotesize{4.5}                  & -               & -                                     &   69.5      &     58.8        & 56.0      & 57.8  & 79.7  &   90.1  \\
Splicing Connector              & 73.4$\pm$\footnotesize{3.2}             & 76.1${\pm}$\footnotesize{2.1}                & -               & -                                     &   64.5      &     59.9        & 89.7      & 79.2  & 84.4  &   81.8  \\ \midrule
\cellcolor[HTML]{FFFFFF}Average & $\mathbf{86.0}\ (\textbf{\textcolor{red}{18.1}}\% \uparrow)$                           & $\mathbf{83.7}\ (\textbf{\textcolor{red}{4.6}}\% \uparrow)$               & 67.9 & 79.1  & 64.3 & 59.5 & 86.0 & 74.0  & 87.7  &   79.7  \\ \bottomrule
\end{tabular}
}
\caption{\textbf{Logical Anomaly detection (classification) performance on MVTec LOCO AD (one-shot).} AnomalyMoE\textdagger  is the \gls{sota} few-short logical \gls{ad} algorithm (CVPR 2024 VAND Challenge Winner). PatchCore, GCAD, ComAD, and AST are all full-shot unsupervised methods trained on all images. GCAD and AnomalyMoE are designed to handle logical \gls{ad}. For each category, we conducted five experiments and calculated the average and standard deviation. Values highlighted in \textcolor{red}{red} indicate increased scores compared to AnomalyMoE\textdagger. The evaluation results for WinCLIP were generated using Anomalib.}
\label{tab:logical_ad_on_loco}
\end{table*}

\section{Experiments and Results}

We conduct comprehensive experiments to evaluate the effectiveness of our algorithm using three \gls{sota} \glspl{avlm}: GPT-4o \cite{achiam2023gpt}, LLaVA 1.6 \cite{liu2024improved}, LLaVA 1.5 \cite{liu2024visual}. Detailed information on the deployment and versions of \glspl{avlm} is provided in Appendix A.4.
All experiments are training-free. Our evaluations are performed on two datasets, MVTec AD and MVTec LOCO AD \cite{Bergmann2019MVTecADComprehensive, DBLP:journals/ijcv/BergmannBFSS22}. We perform one-shot experiments using a single training image and compare our results with competing methods, including full-shot approaches trained on all available images.

\subsection{Dataset and Metrics}
The MVTec LOCO AD dataset \cite{DBLP:journals/ijcv/BergmannBFSS22} is a benchmark for detecting logical anomalies in industrial settings. It comprises five categories, each featuring a variety of \glspl{la}, including missing objects, extra objects, mismatches between colors and objects, and other logical inconsistencies. Additionally, we evaluate our model using MVTec AD~\cite{Bergmann2019MVTecADComprehensive} and focus on some categories, such as screw, pill, toothbrush, capsule, and transistor. These categories are often considered difficult and perform significantly less than others due to their logic-related anomaly characteristics \cite{Jeong2023WinCLIPZero/Few, zhouanomalyclip}. We use $F_1$-max and \gls{auroc} as evaluation metrics, consistent with with \gls{sota} and competing methods.

\subsection{Results}

\noindent \textbf{\textit{Can LogicAD detect naive logical anomalies?}} Current vision features-based algorithms, including WinCLIP and AnomalyCLIP, have shown remarkable performance in SA \gls{ad} while facing challenges in several categories, e.g. , capsule, transistor and toothbrush, from MVTec AD~\cite{Jeong2023WinCLIPZero/Few, zhouanomalyclip, Bergmann2019MVTecADComprehensive}. Although MVTec AD is not designed for evaluating the performance of \gls{la} \gls{ad}, these categories share some naive logical-related characteristics, such as missing objects and mislocation~\cite{Jeong2023WinCLIPZero/Few}. Table~\ref{tab:mvtec_logic_object} and Table \ref{tab:mvtec_eval} indicate that the evaluated algorithms exhibit suboptimal performance in these categories. Since LogicAD is explicitly designed to address logical inconsistencies, Table~\ref{tab:mvtec_logic_object} shows that LogicAD surpasses WinCLIP by an impressive 5.6\% in detecting \gls{la} related categories. Moreover, when compared with other \gls{sota} \gls{ad} methods, such as AnomalyCLIP~\cite{zhouanomalyclip} and VAND~\cite{chen2023april} (all of which are fine-tuned with domain-specific datasets), the training-free LogicAD significantly outperforms the VAND algorithm by 15\% and is comparable to AnomalyCLIP~\cite{zhouanomalyclip}. 

\begin{table}[h!]
\resizebox{\columnwidth}{!}{
\begin{tabular}{@{}ccccccc@{}}
\toprule
\multirow{2}{*}{\begin{tabular}[c]{@{}c@{}}MVTec \\ AD \end{tabular} } & \multirow{2}{*}{SA} & \multirow{2}{*}{LA} & \multicolumn{2}{c}{\textbf{w/o} Training} & \multicolumn{2}{c}{\textbf{w/} Training} \\ \cmidrule(l){4-7} 
        &  &  & LogicAD & WinCLIP &\begin{tabular}[c]{@{}c@{}}Anomaly-\\ CLIP\end{tabular}  & VAND \\ \midrule
Texture & \checkmark & \xmark & 96.9    & 98.1    & 98.7        & 96.9 \\ 
Object*  & \checkmark & \checkmark & 86.5    & 80.9    & 89.1        & 71.5 \\ \bottomrule
\end{tabular}
}
\caption{\textbf{One-shot Logical \gls{ad} performance comparison (AUROC).} The term Object* includes four categories related to \gls{la} in the MVTec AD dataset: capsule, pill, transistor, and toothbrush. Carpet, grid, leather, tile, and wood are categorized as texture and mainly contain \gls{sa}.}
\label{tab:mvtec_logic_object}
\end{table}

\begin{table}[h!]
\resizebox{\columnwidth}{!}{
\begin{tabular}{@{}cccccc@{}}
\toprule
$\mathcal{M}_{GCoT}$ & $\mathcal{M}_{ROI}$ & $\mathcal{M}_{FEmbe}$ & $\mathcal{M}_{LR}$ & AUROC & $F_1$-max \\ \midrule
 \xmark &  \xmark   &  \xmark  &   \xmark &  23.4 & 9.5 \\
 \checkmark &  \xmark  &  \xmark  &  \xmark  & 48.5 & 51.3 \\
 \checkmark & \checkmark &  \xmark  &  \xmark  & 60.4 & 65.3 \\
\checkmark & \checkmark & \checkmark & \xmark   & 86.0 & $\mathbf{83.7}$ \\
\checkmark & \checkmark &\xmark   & \checkmark & N/A & $\mathbf{83.3}$ \\ \bottomrule
\end{tabular}
}

\caption{\textbf{Ablation of different modules in LogicAD model on MVTec LOCO AD.} $\mathcal{M}_{GCoT}$, $\mathcal{M}_{ROI}$, $\mathcal{M}_{FEmbe}$, $\mathcal{M}_{LR}$ denote as Guided CoT, Region of Interest, \textit{format embedding}, and \textit{logic reasoner}. With \textit{logic reasoner}, our model can predict abnormal scores of either 0 or 1 based on \gls{atp}, consequently not applicable for AUROC.}
\label{tab:ablation}
\end{table}

\noindent \textbf{\textit{Can LogicAD detect sophisticated logical anomalies?}} Compared with MVTec AD~\cite{Bergmann2019MVTecADComprehensive}, MVTec LOCO AD~\cite{DBLP:journals/ijcv/BergmannBFSS22} is specifically designed to evaluate the performance of logical \gls{ad} with more sophisticated \gls{la}. Table~\ref{tab:logical_ad_on_loco} presents the performance evaluation of LogicAD on the MVTec LOCO AD dataset. Compared to the \gls{sota} few-shot VLM-based algorithm, AnomalyMoE~\cite{admoe}, LogicAD demonstrates superior performance across all metrics, achieving an increase of $\mathbf{18.1}$\% in \gls{auroc} and $\mathbf{4.6}$\% in $F_1$-max score. Even when compared to full-shot methods, such as PatchCore~\cite{roth2022towards} and AST~\cite{paiss2023teaching}, our method outperforms in many categories. Additionally, when compared to other full-shot algorithms with additional global features such as GCAD and ComAD~\cite{DBLP:journals/ijcv/BergmannBFSS22, liu2023component}, LogicAD exhibits highly competitive results. These findings underscore our model's effectiveness and demonstrate that for non-parametric tasks such as logical understanding, LLMs with \glspl{avlm} have a notable advantage over parametric visual memory bank methods such as PatchCore and AST. Using the \textit{logic reasoner} as shown in Table~\ref{tab:ablation}, we achieve an impressive score of 83.3\%, which is only 0.4\% lower than the score achieved with \textit{format embedding} and only in very few cases, \textit{logic reasoner} has different predictions compared to using \textit{format embedding}, as shown Appendix A.5. However, the advantages of using the \textit{logic reasoner} are substantial, namely, it enhances the model's explainability and eliminates the need for manual or dynamic thresholding, which can cause significant issues in real-world scenarios~\cite{gu2024anomalygpt}. Furthermore, we conduct experiments using different \glspl{avlm}, specifically LLaVA 1.5 and LLaVA 1.6~\cite{liu2024visual, liu2024improved}. Although LLaVA 1.5 peforms worse in $F_1$-max score than the \gls{sota} method, LLaVA 1.6 achieves results comparable to the \gls{sota}. However, regarding AUROC, both LLaVA 1.5 and LLaVA 1.6 outperform the \gls{sota} significantly. This indicates that our method will continue to benefit from future advances in \glspl{avlm} research.

\begin{table}[h!]
        \begin{minipage}{0.48\columnwidth}
        \centering
        \resizebox{\columnwidth}{!}{
        \begin{tabular}{@{}cc@{}}
        \toprule
            \begin{tabular}[c]{@{}c@{}}MVTec LOCO AD \\ (only SA)\end{tabular} & AUROC  \\  \midrule
            LogicAD (ours)                                                  &  81.5     \\
            WinCLIP                                                         &  64.6     \\ 
            GCAD                                                            &  80.7     \\
            AST                                                             &  87.7     \\
            PatchCore                                                       &  89.3     \\ \bottomrule
        \end{tabular}
        }
        \caption{\textbf{LogicAD performance on SA in MVTec LOCO AD dataset.}}
        \label{tab:salogicad}
        \end{minipage} \hspace{0.02\columnwidth}
        \begin{minipage}{0.48\columnwidth}
            \centering
            \resizebox{\columnwidth}{!}{
            \begin{tabular}{@{}ccc@{}}
            \toprule
                VLMs     & AUROC & $F_1$-max   \\ \midrule
                GPT-4o   & \textbf{86.0}  & \textbf{83.2} \\
                LLaVA1.5 & 73.3  & 71.0 \\
                LLaVA1.6 & 76.2  & 78.1    \\ \bottomrule
            \end{tabular}
        }
        \caption{\textbf{LogicAD performance with different \glspl{avlm} backbones on MVTec LOCO AD.}}
        \label{tab:diff_vlm_backbones}
        \end{minipage}
\end{table}

\noindent \textbf{\textit{Can LogicAD detect structural anomalies?}} LogicAD employs a Guided CoT-based methodology primarily for detecting logical anomalies. However, by utilizing carefully curated prompts, our algorithm is also capable of identifying structural anomalies. We evaluated our model on two benchmark datasets: MVTec AD and MVTec LOCO AD. On the MVTec LOCO AD dataset, our model outperforms the one-shot method, WinCLIP and the full-shot method GCAD, but is slightly inferior compared to PatchCore and AST based on Table~\ref{tab:salogicad}. When evaluated on the MVTec AD benchmark, we compared our approach with WinCLIP, a baseline model that does not require fine-tuning with domain-specific datasets and can be applied directly. As shown in Table \ref{tab:mvtec_eval} and Appendix A.6, LogicAD achieves highly competitive scores compared to WinCLIP-based methods, demonstrating its effectiveness in detecting \gls{sa}. As illustrated in Table~\ref{tab:mvtec_logic_object}, for texture categories, LogicAD attains an AUROC of 96.9\%, which is marginally lower than the 98.1\% achieved by WinCLIP and the 98.7\% achieved by AnomalyCLIP, but comparable to VAND~\cite{chen2023april}. These results suggest that while LogicAD is highly effective in general \gls{ad}, particularly those related to logical inconsistencies, there is a minor performance gap when compared to leading train-free vision feature-based methods.

\section{Limitations}
Although our method brings a new perspective and achieves remarkable results, it still has some limitations, such as inconsistent results obtained with different \glspl{avlm} and relatively long inference time with an average of a few seconds per image. While methods such as BitNet or \gls{avlm} pruning can accelerate inference time, such optimizations are beyond the scope of this work~\cite{shang2024llava,wang2023bitnet}. Additionally, we observe some of our failing cases, mainly caused by logical inconsistency in Appendix A.3. Lastly, with \gls{gcot}, our prompts still require minor manual text prompt input, but we note that curated prompts can be reused and need to be defined only once per \gls{ad} task. 

\begin{table}[t]
\resizebox{\columnwidth}{!}{
\begin{tabular}{@{}cccccc@{}}
\toprule
\rowcolor[HTML]{FFFFFF}
\multirow{2}{*}{\begin{tabular}[c]{@{}c@{}}MVTec \\ (AD)\end{tabular}} &                                                                     & \multicolumn{2}{c}{LogicAD}   & \multicolumn{2}{c}{WinCLIP} \\ \cmidrule(l){2-6} 
                          & Category                                                            & AUROC         & $F_1$-max             & AUROC            & $F_1$-max \\ \midrule
\multirow{1}{*}{Texture}  
                          & \begin{tabular}[c]{@{}c@{}}Average \end{tabular}           & 96.9          & \textbf{97.3}  & \textbf{98.1}    & 96.7             \\ \midrule
\multirow{11}{*}{Object}  & {\cellcolor[HTML]{EFEFEF}Capsule}                                   & {\cellcolor[HTML]{EFEFEF}\textbf{84.7}} & {\cellcolor[HTML]{EFEFEF}\textbf{92.2}} & {\cellcolor[HTML]{EFEFEF}77.3}             & {\cellcolor[HTML]{EFEFEF}91.5}             \\
                          & {\cellcolor[HTML]{EFEFEF}Pill}                    & {\cellcolor[HTML]{EFEFEF}\textbf{78.4}} & {\cellcolor[HTML]{EFEFEF}\textbf{91.5}}  & {\cellcolor[HTML]{EFEFEF}78.1}         & {\cellcolor[HTML]{EFEFEF}91.2}             \\
                          & {\cellcolor[HTML]{EFEFEF}Transistor}              & {\cellcolor[HTML]{EFEFEF}\textbf{84.4}} & {\cellcolor[HTML]{EFEFEF}\textbf{81.3}}  & {\cellcolor[HTML]{EFEFEF}81.1}         & {\cellcolor[HTML]{EFEFEF}62.6}            \\
                          & {\cellcolor[HTML]{EFEFEF}Toothbrush}              & {\cellcolor[HTML]{EFEFEF}\textbf{90.0}} & {\cellcolor[HTML]{EFEFEF}\textbf{89.9}}  & {\cellcolor[HTML]{EFEFEF}87.1}         & {\cellcolor[HTML]{EFEFEF}88.1}             \\
                          & Zipper                  & \textbf{93.1} & \textbf{92.5}  & 84.3         & 89.8             \\
                          & Screw                                             & \textbf{89.1} & 81.8           & 74.3             & \textbf{87.5}             \\
                          & Hazelnut                                          & \textbf{}{93.5} & \textbf{95.1}  & 92.2         & 89.7             \\
                          & Bottle                                                              & 79.5          & 81.5           & \textbf{98.7}    & \textbf{96.8}    \\
                          & Cable                                                               & 79.4          & 81.2           & \textbf{85.9}    & \textbf{85.1}   \\
                          & Metal Nut                                                           & 89.6          & 90.1           & \textbf{92.2}    & \textbf{93.2}    \\ \midrule
                          & Objects                                                            & 86.2          & 87.7           & 85.1             & 88.7             \\ \midrule
                          & \begin{tabular}[c]{@{}c@{}}Average\end{tabular} & \textbf{89.7} & 90.9           & 88.9             & \textbf{91.4}    \\ \bottomrule
\end{tabular}
}
\caption{\textbf{Anomaly classification performance comparison on MVTec AD between LogicAD and WinCLIP.} Object categories (the grey-out sections), such as capsule, pill, transistor and toothbrush, contain \glspl{la}, and LogicAD performs better in these categories. Bold indicates the best score.}
\label{tab:mvtec_eval}
\end{table}

\section{Conclusion}

In this paper, we propose a novel framework for \gls{ad} utilizing extracted text from \glspl{avlm}. By incorporating \gls{gcot}, \gls{roi}, and text formatting, our approach leverages the robust logical understanding capabilities of \glspl{avlm}, achieving remarkable one-shot performance in logical \gls{ad} and surpassing \gls{sota} by a significant margin on the latest logical \gls{ad} benchmarks. LogicAD also integrates a theorem prover to predict logical anomalies with corresponding explanations, thereby enhancing the explainability of the model. Our work explores a novel direction in \gls{ad}, demonstrating that using text features can be highly effective, particularly in logical \gls{ad}. In the future, we plan to extend our work by developing a fully automated prompting process by fine-tuning and distilling \glspl{avlm} with logic-related data to reduce inference time while enhancing logical understanding.

\section{Acknowledgments}
This paper is funded by the Deutsche Forschungsgemeinschaft (DFG, German Research Foundation) – 2236/2 and the EU ICT-48 2020 project TAILOR (No. 952215), the German Federal Ministry of Education and Research (BMBF) under the project WestAI (Grant no. 01IS22094D) and  Bio4Monitoring (Grant no. 031B1155).
\bibliography{main}

\end{document}

%% file: glossary_entries.tex
\newacronym{ai}{AI}{artificial intelligence}
\newacronym{xai}{XAI}{explainable \gls{ai}}
\newacronym{ad}{AD}{anomaly detection}
\newacronym{as}{AS}{anomaly segmentation}
\newacronym{dl}{DL}{deep learning}
\newacronym{ml}{ML}{machine learning}
\newacronym{ood}{OOD}{out-of-distribution}
\newacronym{qc}{QC}{quality control}
\newacronym{kd}{KD}{knowledge distillation}
\newacronym{rl}{RL}{reinforcement learning}
\newacronym{gan}{GAN}{generative adversarial network}
\newacronym{cgan}{cGAN}{conditional \gls{gan}}
\newacronym{ae}{AE}{autoencoder}
\newacronym{vae}{VAE}{variational autoencoder}
\newacronym{aae}{AAE}{adversarial autoencoder}
\newacronym{gmm}{GMM}{Gaussian mixture model}
\newacronym{ccd}{CCD}{charge-coupled device}
\newacronym{roi}{ROI}{region of interest}
\newacronym{svm}{SVM}{support vector machine}
\newacronym{smo}{SMO}{sequential minimal optimization}
\newacronym{fpn}{FPN}{feature pyramid networks}
\newacronym{zs}{ZS}{zero-shot}
\newacronym{fs}{few-shot}{}
\newacronym{ov}{OV}{open vocabulary}
\newacronym{vlm}{VLM}{vision-language model}
\newacronym{clip}{CLIP}{contrastive language-image pretraining}
\newacronym{csa}{CSA}{correlative self-attention}
\newacronym{mhsa}{MHSA}{multi-head attention mechanism}
\newacronym{oclr}{OCLR}{one cycle learning rate}
\newacronym{ffn}{FFN}{feed forward network}
\newacronym{roc}{ROC}{receiver operating characteristic}
\newacronym{pr}{PR}{precision-recall}
\newacronym{auroc}{AUROC}{Area Under the Receiver Operating Characteristic}
\newacronym{aupr}{AU\glsentryshort{pr}}{area under the \gls{pr} curve}
\newacronym{auprosavg}{AU\glsentryshort{prosavg}}{area under the \gls{prosavg} curve} 
\newacronym{aupros}{AU\glsentryshort{pros}}{area under the \gls{pros} curve}
\newacronym{prosavg}{\glsentryshort{pr}\textsubscript{os,avg}}{class-averaged, \gls{pros}}
\newacronym{pros}{\glsentryshort{pr}\textsubscript{os}}{open set \gls{pr}}
\newacronym{iou}{IoU}{intersection over union}
\newacronym{tp}{TP}{true positive}
\newacronym{tn}{TN}{true negative}
\newacronym{fp}{FP}{false positive}
\newacronym{fn}{FN}{false negative}
\newacronym{tpr}{TPR}{true positive rate}
\newacronym{tnr}{TNR}{true negative rate}
\newacronym{fnr}{FNR}{false negative rate}
\newacronym{fpr}{FPR}{false positive rate}
\newacronym{pro}{PRO}{per-region-overlap}
\newacronym{flop}{FLOP}{floating-point operation}
\newacronym{fps}{FPS}{frames per second} 
\newacronym{cca}{CCA}{connected component analysis} 
\newacronym{lpips}{LPIPS}{learned perceptual image patch similarity}
\newacronym{nst}{NST}{neural style transfer}

\newacronym{pdf}{PDF}{probability density function}
\newacronym{cdf}{CDF}{cumulative distribution function}
\newacronym{gde}{GDE}{Gaussian density estimation}

\newacronym{mse}{MSE}{mean squared error}

\newacronym{cnn}{CNN}{convolutional neural network}
\newacronym{pca}{PCA}{principal component analysis}  
\newacronym{fc}{FC}{fully connected}
\newacronym{relu}{ReLU}{rectified linear unit}
\newacronym{serlu}{SERLU}{scaled exponentially-regularized linear unit}
\newacronym{oas}{OAS}{oracle approximating shrinkage}  
\newacronym{npca}{NPCA}{negated \gls{pca}}
\newacronym{sed}{SED}{standardized Euclidean distance}
\newacronym{ann}{ANN}{artificial neural network}
\newacronym{mlp}{MLP}{multi-layer perceptron}
\newacronym{nlp}{NLP}{natural language processing}

\newacronym{gda}{GDA}{Gaussian discriminant analysis}
\newacronym{nn}{NN}{nearest neighbor}
\newacronym{1nn}{1-\glsentryshort{nn}}{1-\glsentrylong{nn}}
\newacronym{ocsvm}{oc-SVM}{one-class \gls{svm}}
\newacronym{svdd}{SVDD}{support vector data description}
\newacronym{if}{IF}{isolation forest}
\newacronym{lof}{LOF}{local outlier factor}
\newacronym{odin}{ODIN}{outlier detection using in-degree number}
\newacronym{osr}{OSR}{open set recognition}
\newacronym[longplural={known known classes}]{kkc}{KKC}{known known class}
\newacronym[longplural={known unknown classes}]{kuc}{KUC}{known unknown class}
\newacronym[longplural={unknown unknown classes}]{uuc}{UUC}{unknown unknown class}
\newacronym[longplural={unknown known classes}]{ukc}{UKC}{unknown known class}
\newacronym{msp}{MSP}{maximum softmax probability}
\newacronym{oe}{OE}{outlier exposure}
\newacronym{vrm}{VRM}{vicinal risk minimization}
\newacronym{ebs}{EBS}{energy-based score}
\newacronym{nll}{NLL}{negative log-likelihood}
\newacronym{em}{EM}{expectation maximization}
\newacronym{bic}{BIC}{Bayesian information criterion}
\newacronym{sem}{SCEM}{scanning electron microscopy}
\newacronym{maxlogit}{MaxL}{maximum of the unnormalized predictions} 
\newacronym{fssd}{FSSD}{feature space singularity distance}
\newacronym{maha}{Maha}{Mahalanobis}
\newacronym{db}{d\&b}{dyed \& bleached}
\newacronym{avi}{AVI}{automated visual inspection}
\newacronym{avis}{AVIS}{\gls{avi} system}
\newacronym{aif}{AiF}{German  Federation  of Industrial  Research  Associations}
\newacronym{en}{EN}{EfficientNet}
\newacronym{ci}{CI}{confidence interval}
\newacronym{ewc}{EWC}{elastic weight consolidation}
\newacronym{nf}{NF}{normalizing flows}
\newacronym{vit}{ViT}{vision transformer}
\newacronym{dpi}{DPI}{dots per inch}
\newacronym{bmbf}{BMBF}{German Federal Ministry of Education and Research}
\newacronym{cagr}{CAGR}{compound annual growth rate}
\newacronym{oscr}{OSCR}{open set class-wise recall}
\newacronym{pet}{PET}{polyethylene terephthalate}
\newacronym{olp}{OLP}{OnLoomPattern}
\newacronym{mida}{MIDA}{multiple image defect analysis}
\newacronym{lda}{LDA}{linear discriminant analysis}
\newacronym{shap}{SHAP}{shapley additive explanation}
\newacronym{csd}{CSD}{cosine similarity distance}
\newacronym{pac}{PAC}{probably approximately correct}
\newacronym{ce}{CE}{cross-entropy}
\newacronym{map}{mAP}{mean average-precision}
\newacronym{spade}{SPADE}{semantic pyramid anomaly detection}
\newacronym{pq}{PQ}{panoptic quality}
\newacronym{sq}{SQ}{segmentation quality}
\newacronym{rq}{RQ}{recognition quality}
\newacronym{fcdd}{FCDD}{fully convolutional data descriptor}
\newacronym{padim}{PaDiM}{patch distribution modeling}
\newacronym{avlm}{AVLM}{Autoregressive, multimodal Vision Language Model}
\newacronym{sa}{SA}{structural anomaly}
\newacronym{la}{LA}{logical anomaly}
\newacronym{sota}{SOTA}{state-of-the-art}
\newacronym{llms}{LLMs}{large language models}
\newacronym{atp}{ATP}{automated theorem prover}
\newacronym{cot}{CoT}{Chain-of-Thought}
\newacronym{gcot}{Guided CoT}{Guided Chain-of-Thought}
\newacronym{moe}{MoE}{Mixture of Experts}

%% file: main.bbl
\begin{thebibliography}{45}
\providecommand{\natexlab}[1]{#1}

\bibitem[{Achiam et~al.(2023)Achiam, Adler, Agarwal, Ahmad, Akkaya, Aleman, Almeida, Altenschmidt, Altman, Anadkat et~al.}]{achiam2023gpt}
Achiam, J.; Adler, S.; Agarwal, S.; Ahmad, L.; Akkaya, I.; Aleman, F.~L.; Almeida, D.; Altenschmidt, J.; Altman, S.; Anadkat, S.; et~al. 2023.
\newblock GPT-4 Technical Report.
\newblock \emph{arXiv preprint arXiv:2303.08774}.

\bibitem[{Bergmann et~al.(2022)Bergmann, Batzner, Fauser, Sattlegger, and Steger}]{DBLP:journals/ijcv/BergmannBFSS22}
Bergmann, P.; Batzner, K.; Fauser, M.; Sattlegger, D.; and Steger, C. 2022.
\newblock Beyond Dents and Scratches: Logical Constraints in Unsupervised Anomaly Detection and Localization.
\newblock \emph{International Journal of Computer Vision}, 130(4): 947--969.

\bibitem[{Bergmann et~al.(2019)Bergmann, Fauser, Sattlegger, and Steger}]{Bergmann2019MVTecADComprehensive}
Bergmann, P.; Fauser, M.; Sattlegger, D.; and Steger, C. 2019.
\newblock MVTec AD — A Comprehensive Real-World Dataset for Unsupervised Anomaly Detection.
\newblock In \emph{Conference on Computer Vision and Pattern Recognition}, 9592--9600.

\bibitem[{Breunig et~al.(2000)Breunig, Kriegel, Ng, and Sander}]{breunig2000lof}
Breunig, M.~M.; Kriegel, H.-P.; Ng, R.~T.; and Sander, J. 2000.
\newblock LOF: Identifying Density-based Local Outliers.
\newblock In \emph{Proceedings of the 2000 ACM SIGMOD International Conference on Management of Data}, 93--104.

\bibitem[{Caron et~al.(2021)Caron, Touvron, Misra, J{\'e}gou, Mairal, Bojanowski, and Joulin}]{caron2021emerging}
Caron, M.; Touvron, H.; Misra, I.; J{\'e}gou, H.; Mairal, J.; Bojanowski, P.; and Joulin, A. 2021.
\newblock Emerging Properties in Self-Supervised Vision Transformers.
\newblock In \emph{Proceedings of the IEEE/CVF International Conference on Computer Vision}, 9650--9660.

\bibitem[{Chen, Han, and Zhang(2023)}]{chen2023april}
Chen, X.; Han, Y.; and Zhang, J. 2023.
\newblock APRIL-GAN: A Zero-/Few-Shot Anomaly Classification and Segmentation Method for CVPR 2023 VAND Workshop Challenge Tracks 1\&2: 1st Place on Zero-shot AD and 4th Place on Few-shot AD.
\newblock \emph{arXiv preprint arXiv:2305.17382}.

\bibitem[{Chen et~al.(2024)Chen, Zhou, Shen, Hong, Sun, Gutfreund, and Gan}]{DBLP:conf/aaai/ChenZSHSGG24}
Chen, Z.; Zhou, Q.; Shen, Y.; Hong, Y.; Sun, Z.; Gutfreund, D.; and Gan, C. 2024.
\newblock Visual Chain-of-Thought Prompting for Knowledge-Based Visual Reasoning.
\newblock In \emph{Proceedings of the AAAI Conference on Artificial Intelligence}, volume~38, 1254--1262.

\bibitem[{Enderton(2001)}]{enderton2001mathematical}
Enderton, H.~B. 2001.
\newblock \emph{A Mathematical Introduction to Logic}.

\bibitem[{Girdhar et~al.(2023)Girdhar, El-Nouby, Liu, Singh, Alwala, Joulin, and Misra}]{girdhar2023imagebind}
Girdhar, R.; El-Nouby, A.; Liu, Z.; Singh, M.; Alwala, K.~V.; Joulin, A.; and Misra, I. 2023.
\newblock ImageBind: One Embedding Space To Bind Them All.
\newblock In \emph{Proceedings of the IEEE/CVF Conference on Computer Vision and Pattern Recognition}, 15180--15190.

\bibitem[{Goyal et~al.(2017)Goyal, Khot, Summers-Stay, Batra, and Parikh}]{goyal2017making}
Goyal, Y.; Khot, T.; Summers-Stay, D.; Batra, D.; and Parikh, D. 2017.
\newblock Making the V in VQA Matter: Elevating the Role of Image Understanding in Visual Question Answering.
\newblock In \emph{Proceedings of the IEEE Conference on Computer Vision and Pattern Recognition}, 6904--6913.

\bibitem[{Gu et~al.(2024{\natexlab{a}})Gu, Zhu, Zhu, Chen, Tang, and Wang}]{gu2024anomalygpt}
Gu, Z.; Zhu, B.; Zhu, G.; Chen, Y.; Tang, M.; and Wang, J. 2024{\natexlab{a}}.
\newblock AnomalyGPT: Detecting industrial anomalies using large vision-language models.
\newblock In \emph{Proceedings of the AAAI Conference on Artificial Intelligence}, volume~38, 1932--1940.

\bibitem[{Gu et~al.(2024{\natexlab{b}})Gu, Zhu, Zhu, Chen, and Wang}]{admoe}
Gu, Z.; Zhu, B.; Zhu, G.; Chen, Y.; and Wang, J. 2024{\natexlab{b}}.
\newblock CVPR, Visual Anomaly and Novelty Detection 2.0 Winner 2024, \url{https://www.hackster.io/contests/openvino2024}, {Accessed: 2024-08-01}.

\bibitem[{Gunjal, Yin, and Bas(2024)}]{DBLP:conf/aaai/GunjalYB24}
Gunjal, A.; Yin, J.; and Bas, E. 2024.
\newblock Detecting and Preventing Hallucinations in Large Vision Language Models.
\newblock In \emph{Proceedings of the AAAI Conference on Artificial Intelligence}, volume~38, 18135--18143.

\bibitem[{He et~al.(2016)He, Zhang, Ren, and Sun}]{he2016deep}
He, K.; Zhang, X.; Ren, S.; and Sun, J. 2016.
\newblock Deep Residual Learning for Image Recognition.
\newblock In \emph{Proceedings of the IEEE Conference on Computer Vision and Pattern Recognition}, 770--778.

\bibitem[{Jeong et~al.(2023)Jeong, Zou, Kim, Zhang, Ravichandran, and Dabeer}]{Jeong2023WinCLIPZero/Few}
Jeong, J.; Zou, Y.; Kim, T.; Zhang, D.; Ravichandran, A.; and Dabeer, O. 2023.
\newblock WinCLIP: Zero-/Few-Shot Anomaly Classification and Segmentation.
\newblock In \emph{Conference on Computer Vision and Pattern Recognition}, 19606--19616.

\bibitem[{Kim et~al.(2024)Kim, An, Chikontwe, Kang, Adeli, Pohl, and Park}]{kim2024few}
Kim, S.; An, S.; Chikontwe, P.; Kang, M.; Adeli, E.; Pohl, K.~M.; and Park, S.~H. 2024.
\newblock Few Shot Part Segmentation Reveals Compositional Logic for Industrial Anomaly Detection.
\newblock In \emph{Proceedings of the AAAI Conference on Artificial Intelligence}, volume~38, 8591--8599.

\bibitem[{Lee(2023)}]{lee2023mathematical}
Lee, M. 2023.
\newblock A Mathematical Investigation of Hallucination and Creativity in GPT Models.
\newblock \emph{Mathematics}, 11(10): 2320.

\bibitem[{Li et~al.(2023)Li, Li, Savarese, and Hoi}]{li2023blip}
Li, J.; Li, D.; Savarese, S.; and Hoi, S. 2023.
\newblock BLIP-2: Bootstrapping Language-Image Pre-training with Frozen Image Encoders and Large Language Models.
\newblock In \emph{International Conference on Machine Learning}, 19730--19742.

\bibitem[{Liu et~al.(2024{\natexlab{a}})Liu, Li, Li, and Lee}]{liu2024visual}
Liu, H.; Li, C.; Li, Y.; and Lee, Y.~J. 2024{\natexlab{a}}.
\newblock Improved Baselines with Visual Instruction Tuning.
\newblock In \emph{Proceedings of the IEEE/CVF Conference on Computer Vision and Pattern Recognition}, 26296--26306.

\bibitem[{Liu et~al.(2024{\natexlab{b}})Liu, Li, Li, Li, Zhang, Shen, and Lee}]{liu2024improved}
Liu, H.; Li, C.; Li, Y.; Li, B.; Zhang, Y.; Shen, S.; and Lee, Y.~J. 2024{\natexlab{b}}.
\newblock LLaVA-NeXT: Improved reasoning, OCR, and world knowledge.

\bibitem[{Liu et~al.(2023{\natexlab{a}})Liu, Zeng, Ren, Li, Zhang, Yang, Li, Yang, Su, Zhu et~al.}]{liu2023grounding}
Liu, S.; Zeng, Z.; Ren, T.; Li, F.; Zhang, H.; Yang, J.; Li, C.; Yang, J.; Su, H.; Zhu, J.; et~al. 2023{\natexlab{a}}.
\newblock Grounding dino: Marrying dino with grounded pre-training for open-set object detection.
\newblock \emph{arXiv preprint arXiv:2303.05499}.

\bibitem[{Liu et~al.(2023{\natexlab{b}})Liu, Li, Du, Jiang, Jin, Jin, and Zhao}]{liu2023component}
Liu, T.; Li, B.; Du, X.; Jiang, B.; Jin, X.; Jin, L.; and Zhao, Z. 2023{\natexlab{b}}.
\newblock Component-aware anomaly detection framework for adjustable and logical industrial visual inspection.
\newblock \emph{Advanced Engineering Informatics}, 58: 102161.

\bibitem[{Lu et~al.(2022)Lu, Mishra, Xia, Qiu, Chang, Zhu, Tafjord, Clark, and Kalyan}]{lu2022learn}
Lu, P.; Mishra, S.; Xia, T.; Qiu, L.; Chang, K.-W.; Zhu, S.-C.; Tafjord, O.; Clark, P.; and Kalyan, A. 2022.
\newblock Learn to Explain: Multimodal Reasoning via Thought Chains for Science Question Answering.
\newblock \emph{Advances in Neural Information Processing Systems}, 35: 2507--2521.

\bibitem[{Olausson et~al.(2023)Olausson, Gu, Lipkin, Zhang, Solar-Lezama, Tenenbaum, and Levy}]{olausson2023linc}
Olausson, T.~X.; Gu, A.; Lipkin, B.; Zhang, C.~E.; Solar-Lezama, A.; Tenenbaum, J.~B.; and Levy, R.~P. 2023.
\newblock {LINC}: A Neurosymbolic Approach for Logical Reasoning by Combining Language Models with First-Order Logic Provers.
\newblock In \emph{The 2023 Conference on Empirical Methods in Natural Language Processing}.

\bibitem[{Paiss et~al.(2023)Paiss, Ephrat, Tov, Zada, Mosseri, Irani, and Dekel}]{paiss2023teaching}
Paiss, R.; Ephrat, A.; Tov, O.; Zada, S.; Mosseri, I.; Irani, M.; and Dekel, T. 2023.
\newblock Teaching CLIP to Count to Ten.
\newblock In \emph{Proceedings of the IEEE/CVF International Conference on Computer Vision}, 3170--3180.

\bibitem[{Pan et~al.(2023)Pan, Albalak, Wang, and Wang}]{pan-etal-2023-logic}
Pan, L.; Albalak, A.; Wang, X.; and Wang, W. 2023.
\newblock Logic-{LM}: Empowering Large Language Models with Symbolic Solvers for Faithful Logical Reasoning.
\newblock In Bouamor, H.; Pino, J.; and Bali, K., eds., \emph{Findings of the Association for Computational Linguistics: EMNLP 2023}, 3806--3824. Singapore: Association for Computational Linguistics.

\bibitem[{Radford et~al.(2021)Radford, Kim, Hallacy, Ramesh, Goh, Agarwal, Sastry, Askell, Mishkin, Clark et~al.}]{radford2021learning}
Radford, A.; Kim, J.~W.; Hallacy, C.; Ramesh, A.; Goh, G.; Agarwal, S.; Sastry, G.; Askell, A.; Mishkin, P.; Clark, J.; et~al. 2021.
\newblock Learning Transferable Visual Models From Natural Language Supervision.
\newblock In \emph{International Conference on Machine Learning}, 8748--8763.

\bibitem[{Reiter(1980)}]{reiter1980equality}
Reiter, R. 1980.
\newblock Equality and domain closure in first-order databases.
\newblock \emph{Journal of the ACM (JACM)}, 27(2): 235--249.

\bibitem[{Rippel et~al.(2021)Rippel, Mertens, K{\"o}nig, and Merhof}]{rippel2021gaussian}
Rippel, O.; Mertens, P.; K{\"o}nig, E.; and Merhof, D. 2021.
\newblock Modeling the Distribution of Normal Data in Pre-Trained Deep Features for Anomaly Detection.
\newblock \emph{IEEE Transactions on Instrumentation and Measurement}, 70: 1--13.

\bibitem[{Rippel, Mertens, and Merhof(2021)}]{olivermdnpd}
Rippel, O.; Mertens, P.; and Merhof, D. 2021.
\newblock Modeling the Distribution of Normal Data in Pre-Trained Deep Features for Anomaly Detection.
\newblock In \emph{2020 25th International Conference on Pattern Recognition (ICPR)}, 6726--6733.

\bibitem[{Roth et~al.(2022)Roth, Pemula, Zepeda, Sch{\"o}lkopf, Brox, and Gehler}]{roth2022towards}
Roth, K.; Pemula, L.; Zepeda, J.; Sch{\"o}lkopf, B.; Brox, T.; and Gehler, P. 2022.
\newblock Towards Total Recall in Industrial Anomaly Detection.
\newblock In \emph{Proceedings of the IEEE/CVF Conference on Computer Vision and Pattern Recognition}, 14318--14328.

\bibitem[{Rudolph et~al.(2023)Rudolph, Wehrbein, Rosenhahn, and Wandt}]{rudolph2023asymmetric}
Rudolph, M.; Wehrbein, T.; Rosenhahn, B.; and Wandt, B. 2023.
\newblock Asymmetric Student-Teacher Networks for Industrial Anomaly Detection.
\newblock In \emph{Proceedings of the IEEE/CVF Winter Conference on Applications of Computer Vision}, 2592--2602.

\bibitem[{Shang et~al.(2024)Shang, Cai, Xu, Lee, and Yan}]{shang2024llava}
Shang, Y.; Cai, M.; Xu, B.; Lee, Y.~J.; and Yan, Y. 2024.
\newblock Llava-prumerge: Adaptive token reduction for efficient large multimodal models.
\newblock \emph{arXiv preprint arXiv:2403.15388}.

\bibitem[{Song et~al.(2024)Song, Wang, Li, and Lin}]{song2024good}
Song, Y.; Wang, G.; Li, S.; and Lin, B.~Y. 2024.
\newblock The Good, The Bad, and The Greedy: Evaluation of LLMs Should Not Ignore Non-Determinism.
\newblock \emph{arXiv preprint arXiv:2407.10457}.

\bibitem[{Sun et~al.(2023)Sun, Fang, Wu, Wang, and Cao}]{sun2023eva}
Sun, Q.; Fang, Y.; Wu, L.; Wang, X.; and Cao, Y. 2023.
\newblock EVA-CLIP: Improved Training Techniques for CLIP at Scale.
\newblock \emph{arXiv preprint arXiv:2303.15389}.

\bibitem[{Sur{\'\i}s, Menon, and Vondrick(2023)}]{suris2023vipergpt}
Sur{\'\i}s, D.; Menon, S.; and Vondrick, C. 2023.
\newblock Vipergpt: Visual inference via python execution for reasoning.
\newblock In \emph{Proceedings of the IEEE/CVF International Conference on Computer Vision}, 11888--11898.

\bibitem[{Touvron et~al.(2023)Touvron, Lavril, Izacard, Martinet, Lachaux, Lacroix, Rozi{\`e}re, Goyal, Hambro, Azhar et~al.}]{touvron2023llama}
Touvron, H.; Lavril, T.; Izacard, G.; Martinet, X.; Lachaux, M.-A.; Lacroix, T.; Rozi{\`e}re, B.; Goyal, N.; Hambro, E.; Azhar, F.; et~al. 2023.
\newblock LLaMA: Open and Efficient Foundation Language Models.
\newblock \emph{arXiv preprint arXiv:2302.13971}.

\bibitem[{Wang et~al.(2023)Wang, Ma, Dong, Huang, Wang, Ma, Yang, Wang, Wu, and Wei}]{wang2023bitnet}
Wang, H.; Ma, S.; Dong, L.; Huang, S.; Wang, H.; Ma, L.; Yang, F.; Wang, R.; Wu, Y.; and Wei, F. 2023.
\newblock Bitnet: Scaling 1-bit transformers for large language models.
\newblock \emph{arXiv preprint arXiv:2310.11453}.

\bibitem[{Wang et~al.(2024)Wang, Chen, Chen, Wu, Zhu, Zeng, Luo, Lu, Zhou, Qiao et~al.}]{wang2024visionllm}
Wang, W.; Chen, Z.; Chen, X.; Wu, J.; Zhu, X.; Zeng, G.; Luo, P.; Lu, T.; Zhou, J.; Qiao, Y.; et~al. 2024.
\newblock Visionllm: Large language model is also an open-ended decoder for vision-centric tasks.
\newblock \emph{Advances in Neural Information Processing Systems}, 36.

\bibitem[{Wei et~al.(2022)Wei, Wang, Schuurmans, Bosma, Xia, Chi, Le, Zhou et~al.}]{wei2022chain}
Wei, J.; Wang, X.; Schuurmans, D.; Bosma, M.; Xia, F.; Chi, E.; Le, Q.~V.; Zhou, D.; et~al. 2022.
\newblock Chain-of-Thought Prompting Elicits Reasoning in Large Language Models.
\newblock \emph{Advances in Neural Information Processing Systems}, 35: 24824--24837.

\bibitem[{Ye et~al.(2024)Ye, Chen, Dillig, and Durrett}]{ye2024satlm}
Ye, X.; Chen, Q.; Dillig, I.; and Durrett, G. 2024.
\newblock SatLM: Satisfiability-Aided Language Models Using Declarative Prompting.
\newblock \emph{Advances in Neural Information Processing Systems}, 36.

\bibitem[{Zhang and Wang(2024)}]{zhang2024good}
Zhang, C.; and Wang, S. 2024.
\newblock Good at Captioning, Bad at Counting: Benchmarking GPT-4v on Earth Observation data.
\newblock \emph{arXiv preprint arXiv:2401.17600}.

\bibitem[{Zhang et~al.(2024)Zhang, Cao, Xu, and Shen}]{zhang2024logicode}
Zhang, Y.; Cao, Y.; Xu, X.; and Shen, W. 2024.
\newblock LogiCode: an LLM-Driven Framework for Logical Anomaly Detection.
\newblock \emph{arXiv preprint arXiv:2406.04687}.

\bibitem[{Zhou et~al.(2024)Zhou, Pang, Tian, He, and Chen}]{zhouanomalyclip}
Zhou, Q.; Pang, G.; Tian, Y.; He, S.; and Chen, J. 2024.
\newblock AnomalyCLIP: Object-agnostic Prompt Learning for Zero-shot Anomaly Detection.
\newblock In \emph{The Twelfth International Conference on Learning Representations}.

\bibitem[{Zou et~al.(2022)Zou, Jeong, Pemula, Zhang, and Dabeer}]{Zou2022SPotDifferenceSelf}
Zou, Y.; Jeong, J.; Pemula, L.; Zhang, D.; and Dabeer, O. 2022.
\newblock SPot-the-Difference Self-Supervised Pre-training for Anomaly Detection and Segmentation.
\newblock In \emph{European Conference on Computer Vision}, 392--408.

\end{thebibliography}
